\documentclass[a4paper,10pt,twocolumn]{article}
\usepackage[utf8]{inputenc}
\usepackage{dblfloatfix}
\usepackage{multicol,graphicx}
\usepackage{booktabs}
\usepackage{mathptmx}
\usepackage[T1]{fontenc}
\usepackage{textcomp}
\usepackage{url}
\usepackage{hyperref}

\usepackage[T1]{fontenc}
\usepackage[english]{babel}

\usepackage[backend=bibtex,style=ieee,doi=false,isbn=false,url=true,maxnames=6,citestyle=numeric-comp,giveninits=true]{biblatex}

\fontfamily{ptm}\selectfont

\bibliography{references.bib}

\usepackage[font=small]{caption}
\usepackage{amsmath}

\usepackage{nameref}
\makeatletter
\newcommand\setcurrentname[1]{\def\@currentlabelname{#1}}
\makeatother
\newcommand{\mysection}[1]{\vspace{0.4cm} \uppercase{#1}\setcurrentname{\uppercase{#1}}\phantomsection \vspace{0.4cm}}
\newcommand{\mysubsection}[1]{\hspace{10pt}\textit{#1}\setcurrentname{\textit{#1}}\phantomsection}

\begin{document}
	
\setlength{\textfloatsep}{10pt plus 1.0pt minus 2.0pt}	
\setlength{\columnsep}{1cm}


\twocolumn[%
\begin{@twocolumnfalse}
\begin{center}
	{\fontsize{14}{18}\selectfont
        \textbf{\uppercase{Approximate UMAP allows for high-rate online visualization of high-dimensional data streams}}\\}
    \begin{large}
        \vspace{0.6cm}
        Peter Wassenaar\textsuperscript{1$\star$},
        Pierre Guetschel\textsuperscript{1$\star$}, 
        Michael Tangermann\textsuperscript{1}\\
        \vspace{0.6cm}
        \textsuperscript{$\star$}These authors contributed equally to this work.\\
        \textsuperscript{1}Donders Institute for Brain, Cognition and Behaviour,
        Radboud University, Nijmegen, Netherlands\\
       \vspace{0.5cm}
        E-mail: \href{mailto:pierre.guetschel@donders.ru.nl}{pierre.guetschel@donders.ru.nl}
        \vspace{0.4cm}
    \end{large}
\end{center}	
\end{@twocolumnfalse}%
]%

ABSTRACT: 
In the BCI field, introspection and interpretation of brain signals are desired for providing feedback or to guide rapid paradigm prototyping but are challenging due to the high noise level and dimensionality of the signals.  Deep neural networks are often introspected by transforming their learned feature representations into 2- or 3-dimensional subspace visualizations using projection algorithms like Uniform Manifold Approximation and Projection (UMAP)~\cite{mcinnes_umap_2018}.
Unfortunately, these methods are computationally expensive, making the projection of data streams in real-time a non-trivial task.
In this study, we introduce a novel variant of UMAP, called approximate UMAP (aUMAP). It aims at generating rapid projections for real-time introspection. 
To study its suitability for real-time projecting, we benchmark the methods against standard UMAP and its neural network counterpart parametric UMAP \cite{sainburg_parametric_2021}. 
Our results show that approximate UMAP delivers projections that replicate the projection space of standard UMAP while 
decreasing projection speed by an order of magnitude and maintaining the same training time.

\begin{table*}[!t]
\caption{Summary description of the datasets used to evaluate the accuracy of approximate UMAP.}
\label{table:datasets-rundown}
\begin{tabular*}{\linewidth}{l @{\extracolsep{\fill}} cccl}
\toprule 
Dataset & Classes & Samples total & Dimensionality & Features \\
\midrule
Iris plants & 3 & 150 & 4 & Real positive numbers \\
hand-written digits & 10  & 1797 & 64             & Integers in [0-16]  \\
breast cancer Wisconsin & 2 & 569 & 30 & Real positive numbers \\
\bottomrule
\end{tabular*}
\end{table*}


\mysection{introduction}\label{intro}

The recording of neural signals offers a window into understanding brain activity, with potential applications in various fields. However, a considerable challenge lies in the fact that these signals, particularly electroencephalograms (EEG), are high-dimensional and very susceptible to noise. Consequently, this situation requires the development of specialized analysis techniques to describe and eventually understand the underlying neural processes.

\mysubsection{Introspectability} 
deficiency is an issue for various use cases. An example is the brain-computer interfaces (BCI) field, where providing feedback to the BCI user is key to either improve the BCI's performance \cite{tidoni_audio-visual_2014, sollfrank_effect_2016, luu_gait_2016} or aid in rehabilitation therapies \cite{pillette_why_2020}. Another example is the investigation of novel experimental protocols. To evaluate if a new BCI paradigm is suited, it must be determined if the resulting brain signals contain discriminative information related to the task and if this information is sufficient to accomplish control over an application. Both of these examples could benefit from data introspection in an online environment, as providing feedback immediately may enable a BCI user to adapt on the spot, and experimenters to investigate novel paradigms using a rapid prototyping approach. 

\mysubsection{Interpretability}
challenges of brain data can be tackled by extracting higher-level features, such as the embeddings of a neural network, from the data. 
These features are typically less noisy and have a lower dimensionality, even though they are still too high-dimensional for a human to capture. Obtaining such higher-level features, also known as latent features, is at the core of most machine learning methods in a BCI system. 
For providing feedback to a human and investigating novel paradigms, these features must be visualized. 
This may be done by transforming a set of latent features into a 2D or 3D representation and visualizing these features in an image. This process is known as \textit{projecting}. 

\mysubsection{Projecting}
may be achieved by numerous methods, such as Principal Component Analysis (PCA), Independent Component Analysis (ICA), Uniform Manifold Approximation and Projection (UMAP) \cite{mcinnes_umap_2018}, t-distributed Stochastic Neighbor Embedding (t-SNE) \cite{hinton_visualizing_2008}, and Isometric Mapping (ISOMAP).\\ 
Unfortunately, not all of these methods are well-suited for online projections. For example, while ISOMAP is known to deal well with noisy data, it has a high computational complexity for larger datasets \cite{liu_enhancing_2015}, which increases the model training time and may stagger the projection rate. Alternatively, PCA may be used for rapid projecting, yet it can not account for complex non-linear structure in the data \cite{nyamundanda_probabilistic_2010}. To determine if a projection method is a good fit for online projecting, we defined the following four criteria: 1) The produced projections should be a sufficiently accurate lower dimensional (2D or 3D) representation of the input data. 2) The time it takes to train/fit a model should be relatively brief, i.e., in the range of minutes for the typical data dimensionalities encountered in BCI. 3) Projecting a novel data point into an existing 2D/3D representation should be fast, i.e., take tens to one hundred milliseconds at most, 4) Optimally, the method should be lightweight to avoid strain on the hardware that may impact projection and/or training time, and to avoid requiring specific hardware or technical knowledge to run.

\mysubsection{Uniform Manifold Approximation and Projection (UMAP)}~\cite{mcinnes_umap_2018}  is a good candidate for an online projection. 
It comes with benefits such as utilizing a mathematical model to solve a clearly defined optimization problem, making it lightweight. Additionally, UMAP is widely adopted in the field, as such there are numerous variations of the method. However, UMAP projection times are slow, possibly conflicting with the third criterion. To overcome this potential drawback, parametric UMAP (pUMAP) is a possible alternative \cite{sainburg_parametric_2021}. One of the advantages of pUMAP over standard UMAP is that it generates projections faster, due to utilizing a neural network.  However, pUMAP is less lightweight, which may conflict with criterion~4.

\mysubsection{Approximate UMAP} Motivated by the limitations of existing approaches, we introduce approximate UMAP (aUMAP), a novel alternative that drastically reduces the projection time of UMAP. Its training procedure is identical to standard UMAP and the projection speed increase is achieved by approximating the standard UMAP projections using a nearest neighbors approach.

\mysubsection{Experiments.}
The accuracy of the aUMAP projections, i.e., criterion~1,  is evaluated by comparing them with projections obtained by standard UMAP on three datasets. 
The training and projection times, i.e., criteria~2 and~3, are evaluated for all three UMAP methods. The models will be trained on data and project data characterized by varying dimensionality and sample counts in order to examine the impact of these variables. 

\mysubsection{Structure of the paper}
After introducing details of UMAP, pUMAP and the proposed novel aUMAP method we provide results related to our research questions. In the final discussion section, we debate if the presented UMAP methods satisfy the conditions for online projecting we proposed earlier and provide our conclusion as to which method is most suited for online projecting.

\begin{figure*}[!b]
	\centering
	\includegraphics[scale=.42]{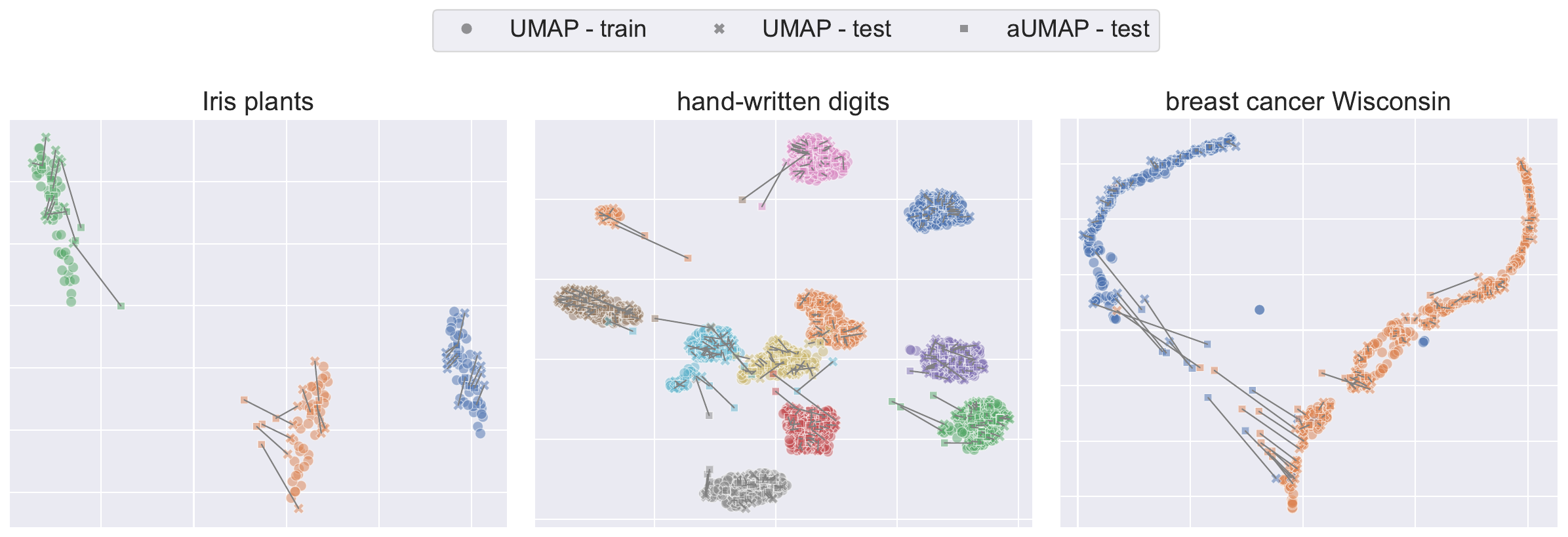}
	\caption{Comparison of UMAP and aUMAP using three datasets. The 2D projections of the training data and test data produced by standard UMAP are displayed in addition to the test set projections produced by aUMAP. The gray lines connect projections of standard UMAP and aUMAP that were obtained from the same test data sample. Colors indicate the classes of the data (not available to the projection methods).}
    \label{performance_projection_space}
\end{figure*}

\mysection{materials and methods}

\mysubsection{Approximate UMAP (aUMAP)}
is designed to be an adaptation of UMAP that reduces the time it takes to produce projections on new data points. It does not deviate from standard UMAP when fitting a model, however, it does train an additional $k$-nearest neighbors model in tandem. Additionally, aUMAP has a different approach for projecting novel data points. It does so by minimizing the summed Euclidian distance between the new projection and the projections of the points that lie closest to the new data point in the input space. The mathematical equation of this is expressed in \autoref{aUMAP}:

\begin{equation}\label{aUMAP}
    u = \sum_{i=1}^{k}\frac{\frac{1}{d_{i}}}{\sum_{j=1}^{k}\frac{1}{d_{j}}}{u_{\underset{}{i}}}
\end{equation}

where $u$ is the projection of a new data point $x$, $k$ is the number of neighbors considered, $u_{1}\dots u_{k}$ are the already existing UMAP projections of the k nearest neighbors $x_{1}\dots x_{k}$ of point $x$ in the input space, and $d_{i} = distance(x,x_{i})$.

\mysubsection{Benchmarking Data:} 
In order to account for the impact of sample size and data dimensionality on training and projection times, we generated a number of mock datasets containing data of varying dimensionality. Each dataset contained an equal number of samples, allowing for the selection of subsets for testing multiple sample counts. The datasets were generated from a multiclass Poisson distribution. These datasets were used for measuring the training and projection times for the varying models.\\
To measure the accuracy of aUMAP, we selected the sklearn datasets Iris plants, hand-written digits, and breast cancer Wisconsin. These datasets were chosen for two reasons. First, they cover a variety of data parameters which includes the number of classes, dimensionality, and sample count, see \autoref{table:datasets-rundown}. Secondly, standard UMAP is able to learn an unsupervised solution for each of these datasets that separates all classes of the data. 

\mysubsection{Model parameters:} 
The models used during the experiments were initiated using the default parameters provided by their base implementation. While this choice clearly leaves room for domain-specific optimizations, we opted to make use of the default parameters to maximize generalizability. The most notable parameters are the distance metric, number of neighbors, minimum distance, and number of components used by  UMAP and nearest neighbors (knn) models. These parameters default to 'euclidean', 15, 0.1, and 2 respectively. For the default parameters of pUMAP see the official documentation \footnote{Paramatric UMAP documentation: \url{https://umap-learn.readthedocs.io/en/latest/parametric_umap.html}}. We diverged from the default parameters on two occasions only. First, we increased the number of nearest neighbors used by the knn model from its default 5 to 15 in order to be consistent with the default of the UMAP model. Secondly, we adapted the parameters of the UMAP and knn models to produce better UMAP projections. For the breast cancer dataset, we set the number of neighbors to 200 and the minimum distance to 1. The minimum distances for the Iris plants and hand-written digits datasets were also increased to 5 and 1, respectively. The remaining parameters were kept the same.

\mysubsection{Approximate UMAP accuracy:} \label{subsec:approx_UMAP_accuracy}
aUMAP only differs from standard UMAP by approximating novel projections instead of calculating them. Both methods seek the same solution. As such, a suitable way to benchmark  aUMAP is to investigate how closely its projections fit those of standard UMAP using the Euclidean distance. The closer the aUMAP projections are in the latent space to their associated standard UMAP projections, the better aUMAP achieves its goal. Here, we refer to associated projections as \textit{latent points} that have been produced using the same input data. We measure the distance as the mean Euclidean distance in the projection space. As the UMAP projections are arbitrarily scaled, the Euclidean distance by itself may not be informative without a normalization by the standard deviation of the projected test points produced by standard UMAP, which we have included for this reason.

\mysubsection{Runtime Measurements:} \label{subsec:runtime_measurements}
Training times were measured for each method across a predefined range of dimensionalities and sample counts. A varying dimensionality was paired with a static sample count, i.e., we selected multiple of the mock datasets, which differed in their data dimensionality, and selected an equal number of training samples across each set. Similarly when varying the sample count, the dimensionality was kept consistent by drawing subsets from a single generated dataset.\\  
When measuring the projection times, the models of the previous step were repurposed to produce the projections, maintaining either a varying dimensionality or number of training samples. A static number of test samples was passed to each model for projecting, matching the data dimensionality of the model's training data. To account for data being presented only a few samples at a time in an online environment, we recorded the training time using two approaches. In the first approach, referred to as \textit{one-go}, all data was given to the model at once, requiring only a singular call to the projection method. The second approach, referred to as \textit{one-go} and designed to better match an online setting, fed data points to each model in small batches of five points at a time. 

\mysubsection{Hardware:} 
All experiments were run using an AMD Ryzen 7 5800x 8-core processor and a NVIDIA GeForce RTX 3060 Ti. Windows Subsystem for Linux (WSL) v.2.0.9.0 was used to enable Tensorflow GPU support. All models, apart from GPU-run pUMAP, were run on CPU.


\mysection{results}\label{results}

\begin{table}[!b]
\caption{Average Euclidean distance between novel standard UMAP and aUMAP projections. Distances are normalized by the standard deviation of the novel standard UMAP projections.}
\label{table:performance_distance}
\begin{tabular*}{\linewidth}{l @{\extracolsep{\fill}} cccl}
\toprule 
Dataset & Mean distance & Variance \\
\midrule
Iris plants & 0.256 & 0.150 \\
hand-written digits & 0.083 & 0.104 \\
breast cancer Wisconsin & 0.126 & 0.211 \\
\bottomrule
\end{tabular*}
\end{table}

\begin{figure*}[!b]
	\centering
	\includegraphics[scale=.42]{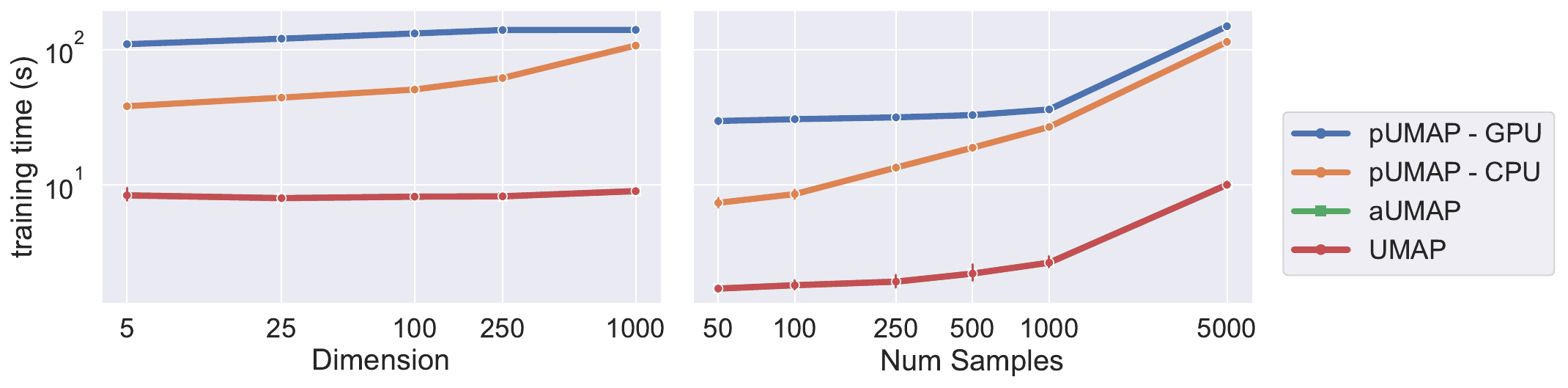}
	\caption{\textbf{Training times}. Models were trained on mock datasets generated from a multiclass Poisson distribution. Left: Models were trained on datasets of 5000 samples with varying dimensionalities. Right: Training across varying sample counts was done using subsets of a 1000-dimensional dataset. Standard UMAP and aUMAP models were trained on the CPU. pUMAP models were trained on both, CPU and GPU separately. Note that aUMAP and standard UMAP results are near-identical, causing the line of the latter to be concealed in the graph. All results shown were averaged across 10 repetitions. Error bars indicate the standard deviation across the runs.}
    \label{training_times}
\end{figure*}

\begin{figure*}[!b]
	\centering
	\includegraphics[scale=.42]{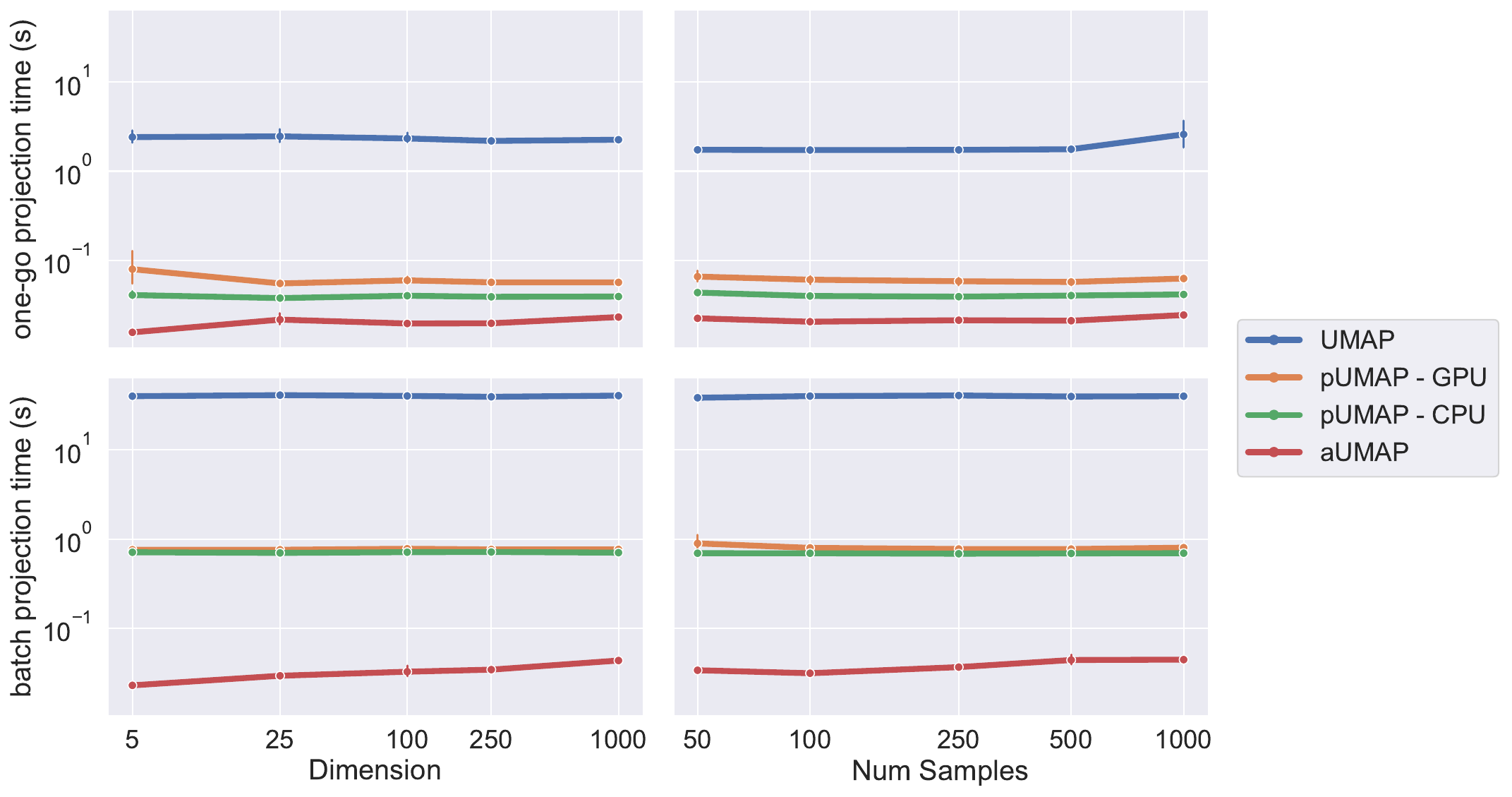}
	\caption{\textbf{Projection times}. The models used for projecting were obtained from the training time experiment. For each model and condition, 500 samples from a multiclass Poisson distribution were passed to the models to be projected. Samples were provided either in a singular batch of 500, denoted as \textit{one-go} (upper figures), or in small batches of 5 samples, denoted as \textit{batch} (bottom figures). Standard UMAP and aUMAP models were trained on CPU. pUMAP was trained on both CPU and GPU separately. The results were averaged across 10 repetitions.}
    \label{projection_times}
\end{figure*}

\mysubsection{Performance of aUMAP:}
\autoref{table:performance_distance} shows the performance of aUMAP given as the average Euclidian distance over all test samples in the dataset in standard deviations between aUMAP projections and their corresponding standard UMAP projection. \autoref{performance_projection_space} allows visualising these projections in addition to the projections of the points used to train the UMAP model used by both methods.
Overall, aUMAP delivers a set of projections that closely match the standard UMAP projections. There is only a small number of aUMAP projections that do not match the classification of the matching standard UMAP projection. 
Across all datasets, the mean distance between the projections lies around 0.1 to 0.25 standard deviations. There is a large variance for each mean distance, which is most extreme for the breast cancer dataset. The variance is reflected by various outliers as shown in \autoref{performance_projection_space}. While the majority of aUMAP projections match the projection space clusters, the occasional projection deviates sharply, sometimes appearing projected closer to a different cluster than its standard UMAP counterpart. Standard UMAP also produces outliers, yet these are fewer and less extreme.

\mysubsection{Training Time:}
\autoref{training_times} visualizes the impact of varying data dimensionality and number of training samples, concerning the time required to train a model. When varying the data dimensionality, the number of training examples used for training was fixed to 5000. When varying the number of training examples, the data dimensionality was fixed to 1000.\\
We observe that the training times of standard UMAP and aUMAP are notably lower than those of pUMAP across any dimensionality or sample count. Training pUMAP on a GPU takes an order of magnitude longer to train than standard UMAP and aUMAP for any dimensionality and sample count. Training pUMAP on a CPU is likewise an order of magnitude slower across all batch sizes and for a dimensionality of 1000, while still being slower than standard UMAP and aUMAP for lower dimensionalities.\\
The data displays an upward trend across all projection methods as the number of samples increases. This trend is most extreme for CPU-run pUMAP and has a minor effect on GPU-run pUMAP training times until increasing the sample count from 1000 to 5000. Only CPU-run pUMAP is strongly affected by an increase in dimensionality.

\mysubsection{Projection Time:}
The projection times are visualized in \autoref{projection_times}. These results were obtained by projecting 500 test samples either in a single batch (\textit{one-go}) or in multiple sub-batches of 5 samples \textit{(batching}). The experiments were repeated 10 times to obtain an average result. Two outlier training times for specific runs were left out in the final averages due to these measurements deviating more than two standard deviations from the average, whereas the remaining projection times were all within one standard deviation. These outliers occurred for CPU-run pUMAP under the \textit{batches} condition, one for dimensionality~=~1000 and the other for sample count~=~1000.\\
The results show that the projection speed of UMAP is significantly worse compared to the other methods. Under the \textit{one-go} condition, standard UMAP takes over a magnitude longer than aUMAP and pUMAP. This difference remains under the \textit{batch} condition between standard UMAP and pUMAP, while increasing by an order of magnitude against aUMAP.\\
The effects of an increase in dimensionality or sample count are only notable for aUMAP in both the \textit{one-go} and \textit{batching} conditions. Other methods do not seem affected, except for standard UMAP under the \textit{one-go} condition at the highest sample count tested, however, this increase is paired with a significant increase in variance.\\
Between the \textit{one-go} and \textit{batch} conditions, aUMAP retains similar projecting times. For standard UMAP and pUMAP, projection times are increased by an order of magnitude when comparing \textit{batch} to \textit{one-go}. As a result, aUMAP runs an order of magnitude faster than pUMAP for the \textit{batched} condition, while running at a similar speed for the \textit{one-go} condition. 


\mysection{Discussion}

In the \nameref{intro}, four requirements are presented to determine if a projection method is suitable for an online setting. 1) The produced projections should be an accurate lower dimensional (2D or 3D) representation of the input data, 2) The time it takes to train/fit a model should be relatively brief, i.e. a timescale of minutes, 3) projections should be rapidly producible in an online fashion on a time scale of tens to 100\, milliseconds, 4) preferably, the method needs to be lightweight to ensure projection or training times do not suffer due to strain on the hardware and to avoid a need for specific hardware or technical knowledge. 

\mysubsection{Approximate UMAP Accuracy:}
Our results in \autoref{performance_projection_space} show that the proposed aUMAP method upholds the clustering produced by standard UMAP. The average distance between the aUMAP projections and standard UMAP projections remains well below one standard deviation of the standard UMAP projections. Although aUMAP produces quantitatively more and more extreme outliers than standard UMAP, it still reproduces the same clustering in the projection space as standard UMAP. This suggests that the accuracy of aUMAP is close enough to that of standard UMAP to suite online projecting, although more prone to outliers. As such, aUMAP satisfies the Criteria~1.

\mysubsection{Training Time:} 
Our results show that all tested methods satisfy the training time condition. The maximum of approx.~2\,minutes was observed for pUMAP when trained on 5000 samples of 1000 dimensions. This time is in line with the requirement we set out. 
However, it should be noted that this method is sensitive to an increase in the number of training samples, so it might violate the requirement if more training data were to be used. 
Unfortunately, we cannot properly assess the runtime of pUMAP when using a GPU. A closer look at the GPU-based pUMAP experiments is presented in a later section. We observe that pUMAP scales with data dimensionality, more so than the other methods, yet to a lesser extent than the input sample size.

Our data shows that GPU-run pUMAP takes more time to train than CPU-run pUMAP. These results go against our prior expectations which expected pUMAP to be trained faster when having access to a GPU. As such, a closer look at GPU-run pUMAP is presented in a later section. This and the following section will focus only on CPU-run pUMAP when discussing pUMAP.\\
Comparing the training speed of CPU-run pUMAP to the other methods, pUMAP is notably outperformed by both standard UMAP and aUMAP, which have about the same training time due to also fitting a UMAP model which is what dominates the training duration. aUMAP fits a nearest neighbors model in addition, yet this only trivially contributes to the training time complexity. Standard UMAP and aUMAP are trained an order of magnitude quicker than CPU-run pUMAP and are less influenced by larger dimensionality and sample size. As a result, standard and aUMAP would each be the best choices for a projection method according to Criteria~2.

\mysubsection{Projection Time:}
The observed projection times display a significant difference between the \textit{one-go} and \textit{batches} condition for all methods except aUMAP, which consistently obtains the lowest projection times. Under the \textit{one-go} condition, the projection time is significantly faster for standard UMAP and CPU-run pUMAP, being an order of magnitude quicker. The projections of aUMAP take longer in the \textit{batch} condition but remains on the same scale.\\
Given that only a few data points are provided at a time in an online environment, the times given in the \textit{batch} condition are of greater interest. Therefore, the following discussion is based on this condition. Standard UMAP requires a significantly longer time to generate projections compared to the other methods. For every dimensionality and training sample count tested, UMAP requires approximately 40\, seconds to project 500 samples, translating to roughly 800\,ms per sample, which is significantly longer than the acceptable projection duration we proposed in our third criterion. As such, standard UMAP cannot be regarded as a good fit for online projecting.  CPU-run pUMAP is a better fit, as it projects the 500 points in just above one second under its slowest conditions, or 2\,ms per data point. This satisfies Criteria~3. aUMAP performs an additional order of magnitude faster, where all 500 projections are generated in less than 100\, milliseconds. Both aUMAP and CPU-run pUMAP scale with the number of training samples and dimensionality, however, not at a significant degree. As such, the impact of the training sample count or data dimensionality on projection time may be disregarded unless dealing with values that are of multiple magnitudes larger than the highest values we tested.\\
This leads to the conclusion that, according to Criteria~3, aUMAP is most suited for online projecting, however, CPU-run pUMAP is also a feasible option.

\mysubsection{GPU-run parametric UMAP:}
Finally, we will discuss GPU-run pUMAP. The results we obtained show that pUMAP runs faster on CPU than on GPU, both when fitting a model and projecting. This is contradictory to our expectations and the results of the paper that introduced pUMAP \cite{mcinnes_umap_2018}. Although the pUMAP study does not comment on the effect of CPU or GPU usage on the model’s training speed, the authors do compare the projection speed of a CPU-run model to that of a GPU-run model, showing that the projection speed is faster for GPU-run pUMAP. According to that paper, GPU-run pUMAP achieves a projection time that is approximately one magnitude lower than CPU-run pUMAP for three of the six datasets used. Additionally, the study shows the effect of data dimensionality on the projection time, comparing 2 to 64 dimensions, which is in line with our results.\\
Based on the results of the original pUMAP paper, we may make an inference on how GPU-run pUMAP would compare to the projection times found for CPU-run pUMAP. At worst, there would be an insignificant difference, at best, GPU-run pUMAP would run an order of magnitude faster. Given the best-case scenario, pUMAP would still project slower than aUMAP. This implies that, given Criteria~3, aUMAP would still be a better projection method for online projecting.\\
Additionally, the pUMAP paper states that the training times of pUMAP are within the same order of magnitude as those of standard UMAP. The study highlights this by showing the cross-entropy loss convergence of pUMAP and standard UMAP, which occurs at around 1 second for two of the three shown datasets and at ${10}^{2}$~seconds for the other dataset. Based on these results, we could speculate that training pUMAP, when having access to a GPU, is as fast as both standard UMAP and aUMAP, making pUMAP an equally valid choice as an online projection method in accordance with Criteria~3.


\mysubsection{Implementation:}
The implementation of the aUMAP algorithm can be found online\footnote{aUMAP implementation: \url{https://neurotechlab.socsci.ru.nl/resources/approx_umap/}}.
Additionally, a graphical application for visualizing incoming data stream in real-time and integrating this algorithm was created in the \footnote{Dareplane: \url{https://github.com/bsdlab/Dareplane}} platform.


\mysection{conclusion}

To conclude, our results suggest that aUMAP can approximate the projection space of standard UMAP sufficiently well for the targeted application. aUMAP may generate projections that lay closer to different clusterings than their standard UMAP counterparts and produces more and more extreme outliers than standard UMAP.\\
Additionally, our results suggest that pUMAP and aUMAP are good fits for real-time projecting. Standard UMAP, on the other hand, does not meet the criteria to be regarded as a good fit due to its projection times being longer than our proposed acceptable maximum. Overall, aUMAP seems the best option for an online projection tool, having the lowest training and projection times while being more accessible than pUMAP. However, aUMAP is more prone to producing outliers in projection space than standard UMAP. As such, if high accuracy is desired pUMAP may be a better choice.

\mysection{Acknowledgements}

We thank Matthias Dold for his insights into the Dareplane platform which was used to implement the graphical application.
Additionally, this work is in part supported by the Donders Center
for Cognition (DCC) and is part of the project Dutch Brain Interface Initiative
(DBI2) with project number 024.005.022 of the research programme Gravitation
which is (partly) financed by the Dutch Research Council (NWO).


\mysection{references}
\printbibliography[heading=none]
\end{document}